\documentclass[runningheads]{llncs}
\usepackage{graphicx}
\newcommand{\orcid}[1]{
	\href{https://orcid.org/#1}{\includegraphics[scale=0.4]{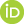}}
}
\usepackage{fontawesome}  
\usepackage{cite}
\usepackage{amsmath,amssymb,amsfonts}

\usepackage[ruled, vlined, linesnumbered]{algorithm2e}

\SetCommentSty{mycommfont}
\usepackage{algpseudocode}
\usepackage{graphicx}
\usepackage{textcomp}
\usepackage{xcolor}
\usepackage{float}
\usepackage{nicefrac}
\usepackage{hyperref}
\usepackage[font=scriptsize]{caption}

\newcommand{\mechanism}{\mathcal{M}}
\newcommand{\universeLog}{\mathcal{L}}
\newcommand{\universeActivity}{\mathcal{A}}

\begin{document}
\title{TraVaG: Differentially Private Trace Variant Generation Using GANs}%\thanks{\scriptsize Funded under the Excellence Strategy of the Federal Government and the L{\"a}nder. We also thank the Alexander von Humboldt Stiftung for supporting our research.}}
\author{Majid Rafiei\orcid{0000-0001-7161-6927}\textsuperscript{\href{mailto:majid.rafiei@pads.rwth-aachen.de}{\faEnvelopeO}} \and
	Frederik Wangelik\orcid{0000-0001-6320-2302} \and
    Mahsa Pourbafrani\orcid{0000-0002-7883-1627} \and
	Wil M.P. van der Aalst\orcid{0000-0002-0955-6940}}
\authorrunning{M. Rafiei et al.}
\institute{Chair of Process and Data Science, RWTH Aachen University, Aachen, Germany \\
% \email{\{majid.rafiei,wvdaalst\}@pads.rwth-aachen.de} 
}
\maketitle              % typeset the header of the contribution
\vspace{-0.1cm}
\begin{abstract}
Process mining is rapidly growing in the industry. Consequently, privacy concerns regarding sensitive and private information included in event data, used by process mining algorithms, are becoming increasingly relevant.
State-of-the-art research mainly focuses on providing privacy guarantees, e.g., differential privacy, for trace variants that are used by the main process mining techniques, e.g., process discovery. 
However, privacy preservation techniques for releasing trace variants still do not fulfill all the requirements of industry-scale usage. Moreover, providing privacy guarantees when there exists a high rate of infrequent trace variants is still a challenge.
In this paper, we introduce TraVaG as a new approach for releasing differentially private trace variants based on \text{Generative Adversarial Networks} (GANs) that provides industry-scale benefits and enhances the level of privacy guarantees when there exists a high ratio of infrequent variants. Moreover, TraVaG overcomes shortcomings of conventional privacy preservation techniques such as bounding the length of variants and introducing fake variants.  
Experimental results on real-life event data show that our approach outperforms state-of-the-art techniques in terms of privacy guarantees, plain data utility preservation, and result utility preservation.

\end{abstract}
\section{Introduction}\label{sec:intro}
Process mining is a family of data-driven techniques for business process discovery, analysis, and improvement. Process mining techniques require event data, which are widely available in most information systems, including ERP, SCM, and CRM systems. During the last decade, process mining has been successfully deployed in many industries, and it has become a crucial success factor for any type of business.
Similar to any data-driven technique in the larger area of data science, concerns about the privacy of people whose data are processed by process mining algorithms are developing as the amount of event data and their usage rise. 
Thus, privacy regulations, e.g., GDPR \cite{gdpr}, restrict data storage and process, which motivates the development of privacy preservation techniques.

Modern privacy preservation methods are mostly based on Differential Privacy (DP), which provides a privacy definition by introducing noise into data.
This is because of its significant properties, including its ability to ensure mathematically proven privacy and protect against PSO (predicate-singling-out) attacks \cite{PSO}.
The purpose of DP-based approaches is to inject noise into the released output in order to conceal the involvement of an individual.
State-of-the-art research in process mining leveraging privacy preservation techniques based on DP focuses on releasing distributions of trace variants, which serve as the foundation for core process mining techniques such as process discovery and conformance checking \cite{processMiningBookWil}.
A trace variant refers to a complete sequence of activities performed for an individual that is considered to be sensitive and private information. In the healthcare context, for instance, a trace variant shows a complete sequence of treatment-related activities performed for a patient that is private information itself and can also be exploited to conclude other sensitive information, e.g., the disease of the patient. 
Table~\ref{tbl:trace_variant} shows a small sample of a trace variant distribution in the healthcare context. Note that in a trace variant distribution, each trace variant is associated with an individual, a so-called case. Moreover, each case has precisely one trace variant.

\begin{table}[t]
\centering
\scriptsize
\caption{A simple event log from the healthcare context, including trace variants and their frequencies.}
\label{tbl:trace_variant}
    \begin{tabular}{l|c}
    \hline
    Trace Variant   & Frequency \\ 
    \hline
    $\langle register, visit, blood\text{-}test, visit, release \rangle$                    & 15 \\ 
    $\langle register, blood\text{-}test, visit, release \rangle$                    & 12 \\
    $\langle register, visit, hospitalization, surgery, release \rangle$                                       & 5 \\ 
    $\langle register, visit, blood\text{-}test, blood\text{-}test, release \rangle$ & 2  \\ \hline
    \end{tabular}
    \vspace{-0.25cm}
\end{table}

% The amount of noise is determined by the sensitivity of the underlying data and privacy parameters $\epsilon$ and $delta$.
% The sensitivity indicates how much uncertainty is required to hide the contribution of one individual to the query. Here, the query is the frequency of each trace variant. Since the modification of only one frequency value, i.e., the contribution of one individual, at a specific release point $i$ in Fig.~\ref{fig:example_traces_original} affects only one trace variant, the sensitivity is set to 1.

To achieve DP for trace variants, conventional so-called \textit{prefix-based} approaches inject noise drawn from a \textit{Laplacian distribution} into the variant distribution obtained from an event log \cite{felix_differential,sacofa}.
These approaches need to generate all possible unique variants based on a set of activities to provide differential privacy for the original distribution of variants. 
Since the set of possible variants that can be generated given a set of activities is infinite, prefix-based techniques need to limit the length of generated sequences. Also, to limit the search space, these approaches typically include a pruning parameter to exclude less frequent prefixes.
Such a process to obtain DP has a high computational complexity and results in the following drawbacks: (1) \textit{introducing fake variants}, (2) \textit{removing frequent true variants}, and (3) \textit{having limited length for generated variants}. 

Several approaches have been proposed to partially or entirely address the aforementioned drawbacks.  
A method, called SaCoFa \cite{sacofa}, aims to mitigate drawbacks (1) and (2) by gaining knowledge regarding the underlying process semantics from the original event data.
However, the privacy quantification of all extra queries to gain knowledge regarding the underlying semantics is not discussed. 
Moreover, the third drawback still remains since this work itself is a prefix-based approach. 
In \cite{mineMe} and a technique called Libra \cite{libra}, which is based on \cite{mineMe}, trace variants are converted to a DAFSA (Deterministic Acyclic Finite State Automata) representation to avoid such drawbacks. However, Libra introduces a clipping parameter for removing infrequent variants. This clipping parameter grows based on the number of unique trace variants and the strength of privacy guarantees. Thus, depending on the number of unique trace variants and privacy parameters, Libra may even remove all the variants and return empty outputs. 
A recent work called TraVaS \cite{rafiei_travas_short} proposes an approach based on \textit{differentially private partition selection strategies} to overcome the above-mentioned drawbacks. Similar to Libra, TraVaS also removes infrequent trace variants. However, in TraVaS, the threshold for removing infrequent variants is only dependent on the input privacy parameters and does not grow with the number of unique variants or the size of event data. Yet, for small event data with a high rate of unique trace variants, TraVaS may not be able to provide strong privacy guarantees.

In this paper, we introduce TraVaG to generate differentially private trace variants from an original variant distribution by means of GANs (Generative Adversarial Networks) \cite{GANs}.
The main idea of TraVaG is to privately learn important event data characteristics. The trained GAN enables the generation of new synthetic anonymized variants that are statistically similar to the original data. Trained generative models work without data access. Thus, as long as the statistical characteristics of the original data do not significantly change, one does not need to apply DP directly to the original event data. For industry-scale big event data, this property can considerably improve the computational complexities \cite{DP_iterative}.
Moreover, TraVaG is based on DP-SGD (Differentially Private - Stochastic Gradient Descent) \cite{DP_SGD} optimization techniques that avoid thresholding on training data or released network outputs. Hence, TraVaG can generate infinite and arbitrarily large anonymized synthetic trace variants even if the original variant frequencies are comparably small.
Moreover, our experiments on real-life event logs demonstrate a better performance of TraVaG compared to state-of-the-art techniques in terms of data utility preservation for the same privacy guarantees. 

The remainder of this paper is structured as follows. In Section~\ref{sec:related_work}, we provide a summary of related work. Preliminaries are provided in Section~\ref{sec:prem}. In Section~\ref{sec:travag}, we present the details of TraVaG. Section~\ref{sec:exp} discusses the experimental results based on real-life event logs, and Section~\ref{sec:conc} concludes the paper.

%\vspace{-2.5 mm}
\section{Related Work}\label{sec:related_work} 
%\vspace{-2.2 mm}
Privacy-preserving process mining is recently growing in importance. 
Several techniques have been proposed to address privacy issues in process mining.
% In this paper, our focus is on the so-called \textit{noise-based} techniques that are based on the notion of \textit{differential privacy}.
In the following, we provide a summary of the work focusing on \textit{releasing differentially private event data} and \textit{generating differentially private event data}.
\vspace{-2 mm}
\subsection{Releasing Differentially Private Event Data}
In \cite{felix_differential}, the authors apply an ($\epsilon, \delta$)-DP mechanism to event logs to privatize \textit{directly-follows relations} and trace variants. 
The underlying principle uses a combination of an ($\epsilon, \delta$)-DP noise generator and an iterative query engine that allows an anonymized publication of trace variants with an upper bound on their length.
In \cite{sacofa}, SaCoFa has been introduced as an extension of \cite{felix_differential}, where the goal is to optimize the query structures with the help of underlying semantics. 
% PRIPEL \cite{pripel} is another extension of \cite{felix_differential}, where more event attributes are integrated into privatized event data.
All the aforementioned techniques follow the so-called prefix-based approach that suffers from the drawbacks explained in Section~\ref{sec:intro}.
To deal with such drawbacks, in \cite{mineMe}, the authors introduced an approach that transforms a trace variant distribution into a DAFSA representation. This approach aims to keep all the original trace variants that may result in high noise injection during the anonymization process. Libra \cite{libra} is a recent work that employs the approach proposed in \cite{mineMe} and aims to increase utility using subsampling and composing privatized subsamples to release differentially private event data. 
TraVaS \cite{rafiei_travas_short} introduces a novel approach based on differentially private partition selection to address the mentioned drawbacks in Section~\ref{sec:intro}.

\subsection{Generating Differentially Private Synthetic Data}
Although DP-based generative Artificial Neural Networks (ANNs) have been quite extensively researched in the major field of data science and machine learning, they have not been used in the context of process mining. Thus, we mainly focus on some of the work outside the domain of process mining.
% In \cite{autoenc2}, the focus is on the challenge of generating mixed-type labeled data with $k$ possible labels. 
% The corresponding algorithm, DP-SYN, first splits the dataset into $k$ labeled subsets and then privately trains an autoencoder on each partition.
In \cite{autoenc1}, the authors adopted a so-called \textit{variational autoencoder}, DP-VAE, which assumes that the mapping from real data to the Gaussian distribution can be efficiently learned.
A different direction was then chosen by \cite{syn3}, where the authors used a \emph{Wasserstein} GAN (WGAN) to generate differentially private mixed-type synthetic outputs employing a Wasserstein-distance-based loss function. 
Finally, in \cite{autogan}, the concepts of WGAN and DP-VAE were combined to first learn a private data encoding and then generate respective encoded data. We adapted this principle for our work to cope with the large dimensionality of event data.

Research in non-private generative models for process mining, primarily focuses on exploiting ANNs and GANs to predict the next state of processes such as \cite{synpro3}, 
% \cite{synpro4}, 
and \cite{synpro2}. Note that the approach in \cite{synpro3} only provides synthetic event data without any privacy guarantees. 

% In the context of non-private generative models for process mining, state-of-the-art research primarily focuses on the activity level of events. Both methods in \cite{synpro3} and \cite{synpro4} train normal GANs to generate new events based on large collections of ground truth. This idea is further specialized in \cite{synpro2} where the authors successfully demonstrated a deep-learning-based framework for repairing missing activity labels.
% Finally, \cite{synpro1} provides a broad comparison of the different existing models and their performance. 

\section{Preliminaries}\label{sec:prem}

% In this section, we introduce the main concepts and definitions utilized throughout the paper. 
We start the preliminaries by introducing basic notations and mathematical concepts.
% The data formats used in the following definitions are strongly linked to basic set operations which will be briefly highlighted.
Let $A$ be a set. $B(A)$ is the set of all multisets over $A$. 
Given $B_1$ and $B_2$ as two multisets, $B_1 \uplus B_2$ is the sum over multisets, e.g., $[a^2,b^3] \uplus [b^2,c^2] = [a^2,b^5,c^2]$.
% A multiset $A$ can be represented as a set of tuples $\{ (a,A(a)) | a \in A \}$ where $A(a)$ is the frequency of $a \in A$.
We define a finite sequence over $A$ of length $n$ as $\sigma {=} \langle a_1, a_2,\dots, a_n\rangle$ where $\sigma(i) {=} a_i {\in} A$ for all $i {\in} \{1,2,\dots,n\}$. The set of all finite sequences over $A$ is denoted with $A^*$. 
% and the set of all elements of $\sigma$ is written as $\{a {\in} \sigma\}$.

\vspace{-0.2cm}
\subsection{\label{subsec:event}Event Log}
Process mining techniques employ event data that are typically collections of unique events recorded per activity execution and characterized by their attributes, e.g., \textit{activity} and \textit{timestamp}. Events in an event log have to be unique.
A \textit{trace} is a single process execution represented as a sequence of events belonging to a case (individual) and having a fixed ordering based on timestamps. 
An event cannot appear in more than one trace or multiple times in one trace.
Our work focuses on the control-flow aspect of an event log that only considers the activity attribute of events in a trace, so-called a \textit{trace variant}. 
 Thus, we define a simple event log based on activity sequences, so-called \textit{trace variants}.

% \begin{definition}[Trace Variant]\label{def:trace_var}
% Let $\universeActivity$ be the universe of activities. A trace variant $\sigma = \langle a_1,a_2,...,a_n \rangle \in \universeActivity^*$ is a sequence of activities performed for a case.
% \end{definition}

\begin{definition}[Simple Event Log]\label{def:simple_el}
\small
Let $\universeActivity$ be the universe of activities. A simple event log $L$ is defined as a multiset of trace variants $\universeActivity^*$, i.e., $L \in B(\universeActivity^*)$. $\universeLog$ denotes the universe of simple event logs.
\end{definition}

In a simple event log representing a distribution of trace variants, one case, which refers to an individual, cannot contribute to more than one trace variant. 

\subsection{\label{subsec:DP}Differential Privacy (DP)}
The main idea of DP is to inject noise into the original data in such a way that an observer who sees the randomized output cannot with certainty tell if the information of a specific individual is included in the data \cite{differential_privacy}.
Considering simple event logs, as our sensitive event data, we define differential privacy in Definition~\ref{def:dp}.

% \begin{definition}[Differential Privacy]\label{def:dp}
% Let $D_1$ and $D_2$ be two neighboring tabular databases that differ only in a single row entry. Further let $\mathcal{M}$ be a randomized mechanism which takes a database as input. Then $\mathcal{M}$ is said to provide $(\epsilon, \delta)$-differential privacy if for all $ S \subseteq rng(\mathcal{M})$, $Pr[\mathcal{M}(D_1) \in S] \leq e^\epsilon \times Pr[\mathcal{M}(D_2) \in S] + \delta$. 
% \end{definition}

% -----Detailed definition----
\begin{definition}[($\epsilon$,$\delta$)-DP for Event Logs]\label{def:dp}
\small
Let $L_1$ and $L_2$ be two neighboring event logs that differ only in a single entry, i.e., $L_2 {=} L_1 {\uplus} [\sigma]$ for any $\sigma {\in} \universeActivity^*$.
Also, let $\epsilon {\in} \mathbb{R}_{>0}$ and $\delta {\in} \mathbb{R}_{>0}$ be two privacy parameters. 
A randomized mechanism $\mechanism_{\epsilon,\delta}{:} \universeLog {\to} \universeLog$ provides ($\epsilon,\delta$)-DP if for all $S {\subseteq} B(\universeActivity^*)$: 
$Pr[\mechanism_{\epsilon,\delta}(L_1) \in S] \leq e^\epsilon {\times} Pr[\mechanism_{\epsilon,\delta}(L_2) \in S] {+} \delta$. 
% Given $L \in \universeLog$, $\mechanism_{\epsilon,\delta}(L) \subseteq \{ (\sigma,L'(\sigma)) \mid \sigma \in \universeActivity^* \wedge L'(\sigma) = L(\sigma) + x_{\sigma} \}$, with $x_{\sigma}$ being realizations of i.i.d. random variables drawn from a probability distribution.
\end{definition}

In Definition~\ref{def:dp}, $\epsilon$ specifies the probability ratio, and $\delta$ allows for a linear violation. In the strict case of $\delta = 0$, $\mechanism$ offers $\epsilon$-DP.
The randomness of respective mechanisms is typically ensured by the noise drawn from a probability distribution that perturbs original variant-frequency tuples and results in non-deterministic outputs. 
The smaller the privacy parameters are set, the more noise is injected into the mechanism outputs, entailing a decreasing likelihood of tracing back the instance existence based on outputs.

% A commonly used $(\epsilon, 0)$-DP mechanism for real-valued statistical queries is the \textit{Laplace} mechanism. This mechanism injects noise based on a Laplacian distribution with scale $\nicefrac{\Delta f}{\epsilon}$.  
% $\Delta f$ is called the sensitivity of a statistical query $f$. Intuitively, $\Delta f$ indicates the amount of uncertainty we must introduce into the output in order to hide the contribution of single instances at $(\epsilon, 0)$-level. In our context, $f$ is the frequency of a trace variant. Since one individual, i.e., a case, contributes to only one trace, $\Delta f {=} 1$. In case an individual can appear in more than one trace, the sensitivity needs to be accordingly increased assuming the same value for the privacy parameter $\epsilon$.
% State-of-the-art event data anonymization frameworks such as our benchmark often use the \textit{Laplace mechanism}.

\subsection{Generative Adversarial Networks (GANs)}
\label{subsec:gan}
A generative adversarial network (GAN) represents a special type of ANN compound to synthesize similar data to its original input. It comprises two separate ANNs, a \textit{generator} and a \textit{discriminator} \cite{GANs}. 
The training principle follows a two-player game: a generator tries to fool the discriminator by generating authentic fake data while a discriminator tries to distinguish between real and fake results. A generator $gen: \mathbb{Z}^m \rightarrow \mathbb{R}^n$ and a discriminator $dis: \mathbb{R}^n \rightarrow \{0,1\}$ can be described as highly parametrizable functions. Here, a generator $gen$ is seeded with random multivariate Gaussian noise $z \in Z^m$ of user-defined dimension $m$ that is translated into a synthetic desired output. A discriminator $dis$ aims to determine whether its input originates from the generator's output. In a simple form, it outputs a binary decision variable, where 0 means the input is fake and 1 means the input is original data. 
In our work, we apply a GAN architecture to synthesize event data.

\subsection{Autoencoder}
\label{subsec:auto}
An \emph{autoencoder} is a certain type of ANN structure used to learn efficient encodings of unlabeled data \cite{autoenc}. The respective encoding is validated and optimized by attempting to regenerate the input from the encoding by decoding.
The autoencoder learns the encoding for a set of data to typically provide dimensionality reduction.
As a result, an autoencoder always consists of two separate ANNs, an encoder $enc: \mathbb{R}^n \rightarrow \mathbb{R}^d$ and a decoder $dec: \mathbb{R}^d \rightarrow \mathbb{R}^n$. These components allow for transforming high-dimensional data $x \in \mathbb{R}^n$ to a compact representation within the so-called latent space $\mathbb{R}^d$ and vice versa (typically $d \ll n$). The specific mappings of $enc$ and $dec$ are characterized by the network's weights and learned from the data during the training phase. 
% In fact, both $enc$ and $dec$ consist of multiple perceptrons stacked into different layers. 
For our work, we employ an autoencoder structure to achieve a compressed encoding of input event data.

\section{TraVaG}\label{sec:travag}
As presented in Section~\ref{sec:related_work}, DP-based generative networks have been extensively researched outside of the process mining context. Typical approaches either adopt variational autoencoder architectures that leverage both encoder and decoder components or GAN architectures employing a discriminator and a generator part. When transferring these ideas to event data, one crucial aspect is the high-dimensional structure that turns out to be challenging during training, particularly if strong DP is added.
Thus, we follow the approach of the novel work \cite{autogan} and \cite{WGAN} that combines the compression functionality of autoencoders with the flexibility of GANs and demonstrated superior performance for general high-dimensional mix-type input data \cite{autogan}. 
Instead of directly generating new event logs, we first learn a compressed encoding and then train a GAN to reproduce data within the encoded latent space. Final datasets are obtained by decoding back the dimension-reduced intermediate format.  
This principle mitigates the complication of GANs when extracting statistical properties from feature-rich data that is limited in size. Particularly, sparse features can be compressed without significant loss of information, while generator networks improve their learning performance due to the lower dimension. Moreover, no Gaussian Mixture distribution is enforced on the latent space, as is the case for typical generative stand-alone autoencoder methods \cite{autoenc1}.
 
 \subsection{The TraVaG Framework}\label{subsec:travag}
\begin{figure}[t]
\centering
\includegraphics[width=.99\textwidth,keepaspectratio]{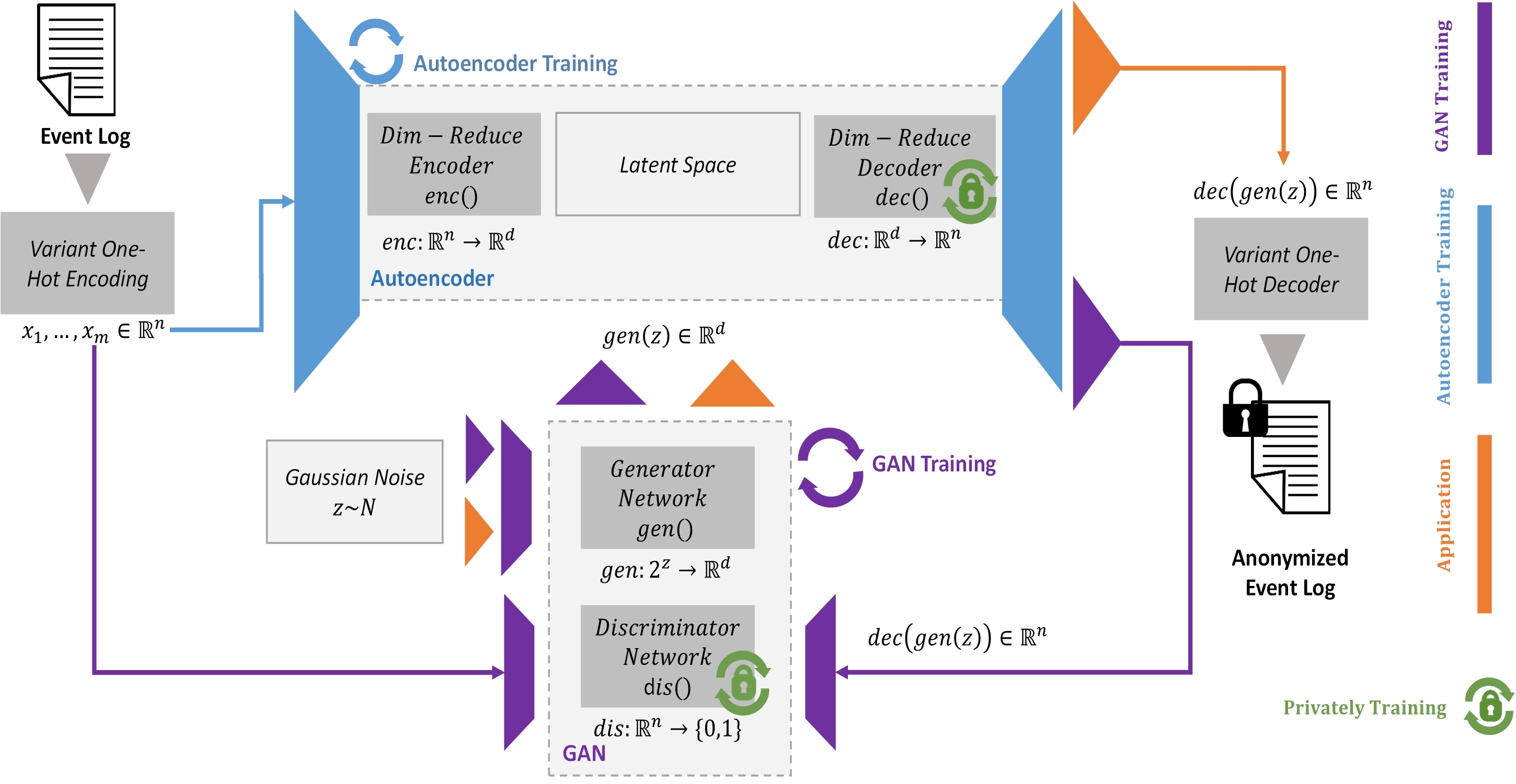}
\caption{A simplified workflow diagram of the TraVaG training and application processes.}
% \caption{Simplified workflow diagram of the TraVaG training and application process. Initially, an event log is transformed into a sparse one-hot-encoded variant table (left upper part). Next, the binary data are fed into the autoencoder procedure (blue) to train both encoder and decoder networks. After the autoencoder training, we use the same data to tune the GAN framework (purple). Here, the generator network takes Gaussian noise $z$ and maps it to the latent space of the autoencoder $G(z)$. The discriminator compares the decoded generated result $D(G(z))$ with the original input data to teach the generator how to construct statistically similar synthetic outputs. Both generator and discriminator are trained together. For the application phase (grey), the generator then again digests Gaussian noise $z$ to produce a compressed data output $G(z)$. We finally decode $G(z)$ and invert the one-hot-encoding to obtain our synthetic new anonymized event log (right lower part). Note that decoder and discriminator are trained and optimized with DP.} 
\label{fig:TraVaG_schema}
\vspace{-0.3cm}
\end{figure}

Different components and the workflow of our framework are shown in Figure~\ref{fig:TraVaG_schema}. We start with preprocessing a simple event log that contains variant distributions in the form of variant-frequency pairs.
There are two common possibilities.
The first option considers the activities within variants and extracts all subsequences of direct neighbors, i.e., Directly-Follows Relations (DFRs). These DFRs are then mapped to a binary or number space and either fed into a GAN as a single feature or as two features along with their frequencies. A downside of this method is that the generator serves as a sequence constructor which allows the creation of artificial variants in the postprocessing phase where all generated activity pairs are linked back together.
To avoid creating fake trace variants, we choose the second option, where only complete variants are considered as inputs. Therefore, a simple event log $L$ with $n$ variants and $m$ cases is binary-encoded as follows.
Within a $m \times n$ matrix, each variant represents a binary feature column and each case denotes a row instance that contains 1 at the respective variant column and 0 elsewhere (sparse matrix). Analogously, this transformation can be inverted back to the original data space.
Thus, TraVaG never produces fake trace variants. Also, one-hot encoding does not influence the data statistics and hence does not incur any privacy costs.
We refer to this preprocessing procedure as \textit{one-hot encoding} and \textit{one-hot decoding}. 
% (\emph{Variant One-Hot-Encoder}, \emph{Variant One-Hot-Decoder} in Figure~\ref{fig:TraVaG_schema}). 
% Since only complete variants are transformed and digested by the network components, no artificial activity combinations can be created.

% We note that the alternative of processing raw event logs, i,e. event data indexed by case id to train generative networks again leads to the issue of fake variants. Here, a GAN would attempt to learn variant information from the attributes of the different cases which naturally represents an imperfect process.

We perform two main training phases including autoencoder training (blue parts) and GANs training (purple parts). 
% In the following, we provide a high-level explanation regarding each training component. 
Since the focus of this work is on the privacy aspect, we describe the privately trained components in more detail. A detailed algorithmic explanation of the training components including the structure of the networks, parameter tuning, activation functions, loss measures, and optimizations is provided in our supplementary document.\footnote{\scriptsize \url{https://github.com/wangelik/TraVaG/blob/main/supplementary/TraVaG.pdf}} 
After the preprocessing, the sparse binary variant vectors $x_1 \dots x_m {\in} \mathbb{R}^n$ are forwarded to the autoencoder training phase, including an encoder and a decoder component. 
These components allow for transforming high-dimensional data $x_i {\in} \mathbb{R}^n$ to a compact representation within the so-called latent space $\mathbb{R}^d$ and vice versa, s.t., $d \ll n$. The dimension $d$ is a hyperparameter of the autoencoder and needs to be selected w.r.t. the GANs configuration. 
Since the encoder does not participate in the process of training the GAN or synthesizing new event data, it does not need to be optimized privately \cite{autodp1,autoenc1}.
The decoder is strongly involved in the anonymization process and is released to the public. Thus, the training of the decoder is performed privately by means of DP-SGD (see Section~\ref{subsec:dp_sgd}). 

% (Differentially Private - Stochastic Gradient Descent)

The same one-hot encoded data $x_1 \dots x_m \in \mathbb{R}^n$ are used to train a GAN consisting of two feed-forward ANNs; a generator $gen: 2^{\mathbb{Z}} \rightarrow \mathbb{R}^d$ and a discriminator $dis: \mathbb{R}^n \rightarrow \{0,1\}$. The goal of the generator $gen$ is to construct synthetic data within the output space $\mathbb{R}^d$ that are similar to the compressed variants. It is seeded with random multivariate Gaussian noise $z$ of a user-defined dimension. The discriminator $dis$ aims at determining whether its input originates from the decompressed generator output $dec(gen(.))$ or from the original data source $x_i, 1 \leq i \leq m$.
Both components are parameterized by their network weights and trained iteratively to outplay each other. Whereas the generator attempts to find latent space outputs that are hard to distinguish from real encoded data by the discriminator, the latter tries to expose these synthetic data records.
Eventually, this principle enables the generator to learn and capture the statistical properties of the input variant distribution through the lenses of the autoencoder.
Note that due to the integrated autoencoder, the generator only targets the latent space $\mathbb{R}^d$ which is much easier to achieve than constructing data in $\mathbb{R}^n$. Also, it averts to access the real confidential data space and does not need to be trained with DP as opposed to the discriminator that is again privately optimized with DP-SGD algorithms \cite{autogan}.

Once both the autoencoder and GAN are trained, one can generate new synthetic anonymized event data (orange parts). The underlying mechanism equals the training step of the generator. Starting with a random Gaussian noise sample $z$, this noise becomes digested by the generator, yielding $gen(z)$. From the latent space, the decoder then maps $gen(z)$ to $dec(gen(z))$. Finally, the synthetic one-hot encoded result is transformed back to the variant universe. 
One compelling advantage of TraVaG lies in the underlying data format. Since the feature space represents the different variants of the original data, TraVaG considers them as given and only has to learn their distribution during training. When applied, the framework reconstructs an anonymized version of this distribution over multiple runs without introducing new variants. 
% As soon as sequence-based data would be used, e.g. raw event logs, this benefit directly disappears.
The more synthetic data are created, the better the consolidated TraVaG output, i.e., new anonymized variants better approximate the original variant distribution. Note that this process does not converge to the true variant frequencies, but to the TraVaG-internal learned anonymous version. Thus, it is recommended to run TraVaG at least as often as the number of cases in the original event log. In case smaller privatized datasets are needed, the output can be down-sampled during postprocessing rounds.

% \subsection{decomposition theorem}
% Theorem~\ref{theorem:comp} demonstrates an important property of $(\epsilon, \delta)$-DP that refers to its behavior under compositional deployment.

% \begin{theorem}[$(\epsilon, \delta)$-DP Mechanism Composition \cite{differential_privacy}]\label{theorem:comp}
% Let $L$ be a simple log, and $\mathcal{M}_i,\allowbreak 1 {\leq} i {\leq} n$ be $(\epsilon_i, \delta_i)$-DP mechanisms. The sequential application of these mechanisms on arbitrary sublogs of $L$ leads to an overall worst-case privacy level parameterized by $(\sum_{1\leq i \leq n} \epsilon_i, \allowbreak \sum_{1\leq i \leq n} \delta_i)$. 
% If each $\mathcal{M}_i$ operates on strictly disjoint sublogs of $L$, the worst-case privacy level is $(max_{1\leq i \leq n}~\epsilon_i, max_{1\leq i \leq n}~\delta_i)$, so-called parallel composition.
% \end{theorem}

% As Theorem~\ref{theorem:comp} states, different $(\epsilon, \delta)$-DP mechanisms can be easily combined into more complex algorithms at the cost of a directly measurable cumulative privacy loss. Nevertheless, the result still promises $(\epsilon, \delta)$-DP independent of the exact form of composition or query structure.

\subsection{Differentially Private - Stochastic Gradient Descent (DP-SGD)}
\label{subsec:dp_sgd}
To render SGD differentially private, Abadi et al. \cite{DP_SGD} proposed the following two steps. 
Given a dataset $X = \{x_i \in \mathbb{R}^n \mid 1 \leq i \leq m\}$, $f$ as a loss function, and $\theta$ as the model parameter. 
First, the gradient $g_i = \nabla_\theta f_\theta(x_i)$ of each data sample $x_i$ is clipped at some real value $C \in \mathbb{R}_{>0}$ to ensure its $L^2$-norm of the gradient does not exceed the clipping value. 
For our work, we refer to the following clipping function\footnote{ \scriptsize Note that also other clipping strategies exist, as highlighted in \cite{DP_iterative}.}: $\text{clip}(g_i,C) = g_i \cdot \min\left(1,C/{||g_i||_2}\right)$. 

% \begin{equation}
%     \text{clip}(x,C):=x \cdot \min\left(1,C/{||x||_2}\right)
% \end{equation}

Then, as Equation~\ref{eq:noise_gradient_single} shows, multivariate Gaussian noise parametrized by a noise multiplier $\Phi \in \mathbb{R}$ is added to the clipped gradient vectors before averaging over the batch $B \subseteq X$. We further denote the identity matrix as $I$ and the Gaussian distribution of unspecified dimension as $\mathcal{N}$.

\begin{equation}\label{eq:noise_gradient_single}
\small
g_B \leftarrow\textstyle\frac{1}{|B|}\left(\sum_{i\in B}\text{clip}(\nabla_\theta f_\theta(x_i),C) 
+\mathcal{N}(0,C^2\Phi^2I)\right)
\end{equation}

The noisy-clipped-averaged gradient $g_B$ is now differentially private and can be used for conventional descent steps: $\theta \leftarrow \theta - \eta \cdot g_B$, where $\eta$ is the so-called \textit{learning rate}.
Note that clipping the individual gradients as in Equation~\ref{eq:noise_gradient_single} can also be replaced by instead clipping gradients of groups of more data points, so-called \emph{microbatches} \cite{DP_iterative}. 
Instead of the common DP parameters $\epsilon$ and $\delta$, DP-SGD uses the related noise multiplier $\Phi$. When translating between these two types of settings, novel research has demonstrated a tighter privacy bound if the batch sampling process for $B$ is conducted according to a specific procedure \cite{DP_SGD}.
This procedure independently selects each data point of $X$ with a fixed probability $q$, the so-called \textit{sampling rate}, in each step.

% Mathematically, the original batch $B$ is therefore further partitioned into new batches $B_1, \dots, B_k \subseteq B$ each of size $r$ (skipping the dividend). We then obtain the new microbatch-related gradient as in Equation~\ref{eq:noise_gradient_mult}.

% \begin{equation}\label{eq:noise_gradient_mult}
% \small
% g_B \leftarrow\frac {1}{k}\left(\textstyle\sum_{i=1 }^k \text{clip}(\nabla_\theta f_\theta(X_{B_i}) 
% ,C)+\mathcal{N}(0,C^2\Phi^2I)\right).  
% \end{equation}

% Naturally, standard differentially private SGD (DP-SGD) corresponds to setting $r=1$. Increasing $r$ while holding $|B|$ fixed primarily decreases the runtime and reduces the achieved training accuracy. Also, it has been shown to not significantly impact privacy for large datasets \cite{DP_iterative}.

% Instead of the common DP parameters $\epsilon$ and $\delta$, DP-SGD uses the related noise multiplier $\Phi$. When translating between these two types of settings, a novel research has demonstrated a tighter privacy bound if the batch sampling process for $B$ is conducted according to a specific procedure \cite{DP_SGD}.
% This procedure independently selects each data point of $X$ with a fixed probability $q$, the so-called \textit{sampling rate}, in each step.

\subsection{Privacy Accounting}
\label{subsec:privacy_acc}

To evaluate the exact privacy guarantee provided by DP-SGD algorithms, we employ the so-called \emph{Renyi Differential Privacy} (RDP) \cite{rdp}, a 
different notion of DP typically used for private optimization. RDP is defined based on the concept of \textit{Renyi divergence}. Given two probability distributions $P$ and $Q$, the Renyi divergence of order $\alpha$ is defined as follows: \small $D_\alpha(P||Q) := \frac{1}{\alpha-1} \log \mathbb{E}_{x\sim Q} \left( \frac{P(x)}{Q(x)} \right)^\alpha$.\normalsize

% \begin{definition}[($\alpha, \epsilon$)-RDP]
% A randomized mechanism $\mathcal{M}$ is said to be $(\alpha, \epsilon)$-RDP if for all neighboring databases $K, K'$ that differ in only one entry

% \begin{equation*}
% RDP(\alpha) := D_\alpha(\mathcal{M}(K)||\mathcal{M}(K'))\leq \epsilon,
% \end{equation*}

% where \(D_\alpha(P||Q) := \frac{1}{\alpha-1} \log \mathbb{E}_{x\sim Q} \left( \frac{P(x)}{Q(x)} \right)^\alpha\) is the \emph{Renyi divergence} of order \(\alpha\) between two distributions \(P\) and \(Q\).
% \end{definition}

\begin{definition}[($\alpha, \epsilon$)-RDP for Event Logs]
\small
Let $L_1$ and $L_2$ be two neighboring event logs that differ only in a single entry, e.g., $L_2 {=} L_1 {\uplus} [\sigma]$ for any $\sigma {\in} \universeActivity^*$.
Given $\alpha > 1$ and $\epsilon \in \mathbb{R}_{>0}$,
a randomized mechanism $\mechanism_{\alpha,\epsilon}{:} \universeLog {\to} \universeLog$ provides $(\alpha,\epsilon)$-RDP if $D_\alpha(\mathcal{M}(L_1)||\mathcal{M}(L_2)) \leq \epsilon$.
\end{definition}

To obtain the final $(\epsilon,\delta)$-DP parameters, we employ the following two propositions on the composition of ($\alpha, \epsilon$)-RDP mechanisms and the conversion of ($\alpha, \epsilon$)-RDP parameters to ($\epsilon, \delta$)-DP parameters. 
% Considering two RDP mechanisms $\mathcal{M}_1$ and $\mathcal{M}_2$, we formulate a composition principle as Proposition~\ref{prep:com_RDP} \cite{rdp}. Moreover, due to the conceptual similarity between $(\alpha, \epsilon)$-RDP and ($\epsilon, \delta$)-DP, the corresponding privacy parameters can be converted as shown in Proposition~\ref{prep:rdp_conver} \cite{rdp}.
\vspace{-0.04cm}

\begin{proposition}[Composition of RDP \cite{rdp}]\label{prep:com_RDP}
\small
If $\mathcal{M}_1$ and $\mathcal{M}_2$ are two $(\alpha, \epsilon_1)$-RDP and $(\alpha,\epsilon_2)$-RDP mechanisms for $\alpha > 1$, respectively. Then, the composition of $\mathcal{M}_1$ and $\mathcal{M}_2$ satisfies $(\alpha,\epsilon_1 +\epsilon_2)$-RDP.\label{prop:rdpcomp}
\end{proposition}

\begin{proposition}[RDP Parameter Conversion \cite{rdp}]\label{prep:rdp_conver}
\small
If a mechanism $\mathcal{M}$ satisfies $(\alpha, \epsilon)$-RDP with $\alpha > 1$, then for all $\delta>0$, $\mathcal{M}$ satisfies $(\epsilon+(\log 1/\delta)/(\alpha-1), \delta)$-DP.\label{prop:rdpconv}
\end{proposition}

During an iterative application of Gaussian mechanisms, as is the case in DP-SGD, the Renyi divergence allows more tightly capturing of the corresponding privacy loss than standard ($\epsilon,\delta$)-DP.
To compute the final $(\epsilon,\delta)$-DP parameters from multiple runs of DP-SGD, the following three steps are followed.
% Based on the sampling strategy, first, a so-called \emph{subsampled Renyi divergence} needs to be derived. Subsequently, privacy is composed under RDP and then converted back into DP.

\begin{enumerate}
%\vspace{-1.5 mm}
    \small
    \item \textbf{Subsampled RDP.} Given a sampling rate $q$ and noise multiplier $\Phi$, the RDP privacy parameters for one iteration of DP-SGD can be derived as a non-explicit integral function of $\alpha \geq 1$ \cite{rdp}. This function is standardized in many privacy-related optimization packages and will be referred to as RDP$_1(q,\Phi)$ \cite{DP_SGD}.
    \item \textbf{RDP Composition.} Since DP-SGD is most likely to run iteratively, we need to compose Step 1 over all executions according to Proposition~\ref{prop:rdpcomp}. Hence, the resulting RDP parameters of $T$ iterations are obtained by computing RDP$_T(q,\Phi,T) := \text{RDP}_1(q,\Phi) \cdot T$.
    \item \textbf{Conversion to $(\epsilon,\delta)$-DP.} After retrieving an expression for the overall RDP privacy parameters with RDP$_T$, we need to convert the respective $(\alpha,\epsilon)$ tuple to a $(\epsilon,\delta)$ guarantee according to Proposition~\ref{prop:rdpconv}.
    Since the $\epsilon$ parameter of RDP is also a function of $\alpha$, Step 3 involves optimizing for $\alpha$ to achieve a minimal $\epsilon$ and $\delta$.
%\vspace{-1.5 mm}
\end{enumerate}

We apply this procedure to obtain the respective privacy guarantees $(\epsilon,\delta)$-DP on both the autoencoder and the GAN-based discriminator of TraVaG. The resulting values are then combined into a final privacy cost by the \textit{composition theorem} of DP  \cite{differential_privacy}.
According to the composition theorem, different $(\epsilon, \delta)$-DP mechanisms can be easily combined into more complex algorithms at the cost of a directly measurable cumulative privacy loss, and the result still promises $(\epsilon, \delta)$-DP independent of the exact form of composition or query structure.

% \begin{theorem}[$(\epsilon, \delta)$-DP Mechanism Composition for Event Logs]\label{theorem:dp_comp}
% Let $L$ be a simple log, and $\mathcal{M}_i,\allowbreak 1 {\leq} i {\leq} n$ be $(\epsilon_i, \delta_i)$-DP mechanisms. The sequential application of these mechanisms on arbitrary sublogs of $L$ leads to an overall worst-case privacy level parameterized by $(\sum_{1\leq i \leq n} \epsilon_i, \allowbreak \sum_{1\leq i \leq n} \delta_i)$. 
% If each $\mathcal{M}_i$ operates on strictly disjoint sublogs of $L$, the worst-case privacy level is $(max_{1\leq i \leq n}~\epsilon_i, max_{1\leq i \leq n}~\delta_i)$, so-called parallel composition.
% \end{theorem}

% As Theorem~\ref{theorem:dp_comp} states, different $(\epsilon, \delta)$-DP mechanisms can be easily combined into more complex algorithms at the cost of a directly measurable cumulative privacy loss. Nevertheless, the result still promises $(\epsilon, \delta)$-DP independent of the exact form of composition or query structure.

\section{Experiments}\label{sec:exp}
We evaluate the performance of TraVaG on real-life event logs.
We select two event logs of varying sizes and trace uniqueness. As we discussed in Section~\ref{sec:intro} and stated in other research such as \cite{felix_differential}, \cite{sacofa},
% \cite{rafiei_group}, 
and \cite{libra} infrequent variants are challenging to privatize. Thus, trace uniqueness is an important analysis criterion.
The Sepsis log describes hospital processes for Sepsis patients and contains many rare traces \cite{sepsis}. 
In contrast, BPIC13 has significantly more cases at a four times smaller trace uniqueness \cite{bpic13}. BPIC13 describes an incident and problem management system called VINST. 
% Finally, the BPIC-2012-App log from \cite{bpic12} describes the processes of different loan applications from a Dutch financial institute.
Both logs are realistic examples of confidential human-centered information where the case identifiers refer to individuals.
Table~\ref{tab:exp_data} shows detailed log statistics.

\begin{table}[b]
\footnotesize
\centering
\caption{General statistics of the event logs used in our experiments.}
\label{tab:exp_data}
\begin{tabular}{c|c|c|c|c|c}
\hline
Event Log & \#Events & \#Cases & \#Activities & \#Variants & Trace Uniqueness\\
\hline
Sepsis & 15214 & 1050 & 16 & 846 & 80\%\\
BPIC13 & 65533 & 7554 & 4 & 1511 & 20\%\\
\hline
\end{tabular}
\vspace{-2 mm}
\end{table}

We perform our evaluation for a wide range of the main privacy parameters $\epsilon {\in} \{0.01, 0.1, 1, 2\}$ and $\delta {\in} \{10^{-6},10^{-5},10^{-4},10^{-3},0.01\}$.
These ranges are selected in accordance with typical values employed at industrial applications as well as state-of-the-art DP research \cite{felix_differential, sacofa, libra, apple}.
We particularly note that extreme settings such as $\epsilon=2, \delta=0.5$ are not chosen due to practical relevance, but to demonstrate how the anonymization methods behave when starting from a weak- or non-private environment.
% \footnote{ \scriptsize In general, $\delta$ is recommended to be not bigger than $1/|D|$, where $|D|$ is the size of dataset $D$ \cite{differential_privacy}.}  
Due to the probabilistic nature of $(\epsilon, \delta)$-DP, we run the TraVaG generator 100 times on all input event logs and all privacy parameters and report the average values.
We compare our results with TraVaS \cite{rafiei_travas_short} as a state-of-the-art technique  and the original prefix-based framework called benchmark \cite{felix_differential}.\footnote{\scriptsize Note that in \cite{rafiei_travas_short}, TraVaS was already compared with SaCoFa \cite{sacofa} and benchmark \cite{felix_differential} and showed better performance. Here, the benchmark method is included for easier comparison. Moreover, Libra \cite{libra} does not take $\epsilon$ as an input parameter but computes it based on $\alpha$ as an RDP parameter and its sampling strategy. This makes the comparison based on exact $\epsilon$ and $\delta$ parameters very difficult. Nevertheless, an important observation in contrast to TraVaG is that Libra returns an empty log for event logs with many infrequent variants, such as Sepsis when $\delta \leq 10^{-3}$.} 
The sequence cutoff for the benchmark method is set to the length that covers 80\% of variants in each log, and the remaining pruning parameter is adjusted such that on average anonymized logs contain a comparable number of variants with the given original log.
The ANNs of TraVaG are configured by a semi-automated tuning approach w.r.t the different input logs.
Whereas most design decisions and hyperparameters are tweaked according to results of manual tests as well as research experience, the settings: \emph{batch size} ($B$), \emph{number of iterations} ($I$) and \emph{noise multiplier} ($\Phi$) are automatically optimized via a grid-search \cite{gridsearch} for fixed privacy levels. A detailed list of the derived settings for each event log, the concrete network designs, and configuration values are available on GitHub.\footnote{\scriptsize \url{https://github.com/wangelik/TraVaG/blob/main/supplementary/TraVaG.pdf}}

 \vspace{-1 mm}
\subsection{Evaluation Measures}\label{subsec:eval_measures}
Suitable evaluation measures are required to assess the performance of an $(\epsilon, \delta)$-DP mechanism in terms of data (result) utility preservation.
The \textit{data utility} perspective measures the similarity between two logs independent of future applications. 
For evaluating data utility we employ the following measures: \textit{relative log similarity} \cite{rafiei_quant,rafiei_travas_short} and \textit{absolute log difference} \cite{rafiei_travas_short}. 
\textit{Relative log similarity} measures the \textit{earth mover's distance} between two trace variant distributions, where the normalized \textit{Levenshtein} string edit distance is used as a similarity function between trace variants.
This measure quantifies the degree to which the variant distribution of an anonymized log matches the original variant distribution on a scale from 0 to 1. 
\textit{Absolute log difference} accounts for the situations where distribution-based measures provide misleading expressiveness \cite{rafiei_travas_short}. Exemplary cases are event logs possessing similar variant distributions, but significantly different sizes. 
% For such scenarios, \emph{relative log similarity} shows high similarity results, whereas \emph{absolute log difference} can detect these size disparities.
To calculate an absolute log difference value, we use the approach introduced in \cite{rafiei_travas_short}, where input logs are converted to a \textit{bipartite graph} of variants as vertices. Then, a \textit{cost network flow} problem is solved by setting demands and supplies to the absolute variant frequencies and utilizing a \textit{Levenshtein} distance between variants as an edge cost. 
Thus, the result of this measure shows the minimal number of \textit{Levenshtein} operations to transform variants of an anonymized log into variants of the original log. 
Details of the exact algorithms are available.\footnote{\scriptsize \url{https://github.com/wangelik/TraVaG/blob/main/supplementary/metrics.pdf}}

We additionally evaluate the performance of TraVaG in terms of \textit{result utility preservation} for \textit{process discovery} as a specific application of trace variant distribution. 
In this respect, we use the \textit{inductive miner infrequent} \cite{sander_inductive} with a default noise threshold of 20\% to discover process models from the privatized event logs for all $(\epsilon, \delta)$ settings under investigation. Then, we compare the models with the original event log to obtain token-based replay \textit{fitness} and \textit{precision} scores \cite{processMiningBookWil}.

\begin{figure}[bt]
\centering
\includegraphics[width=0.92\textwidth,keepaspectratio]{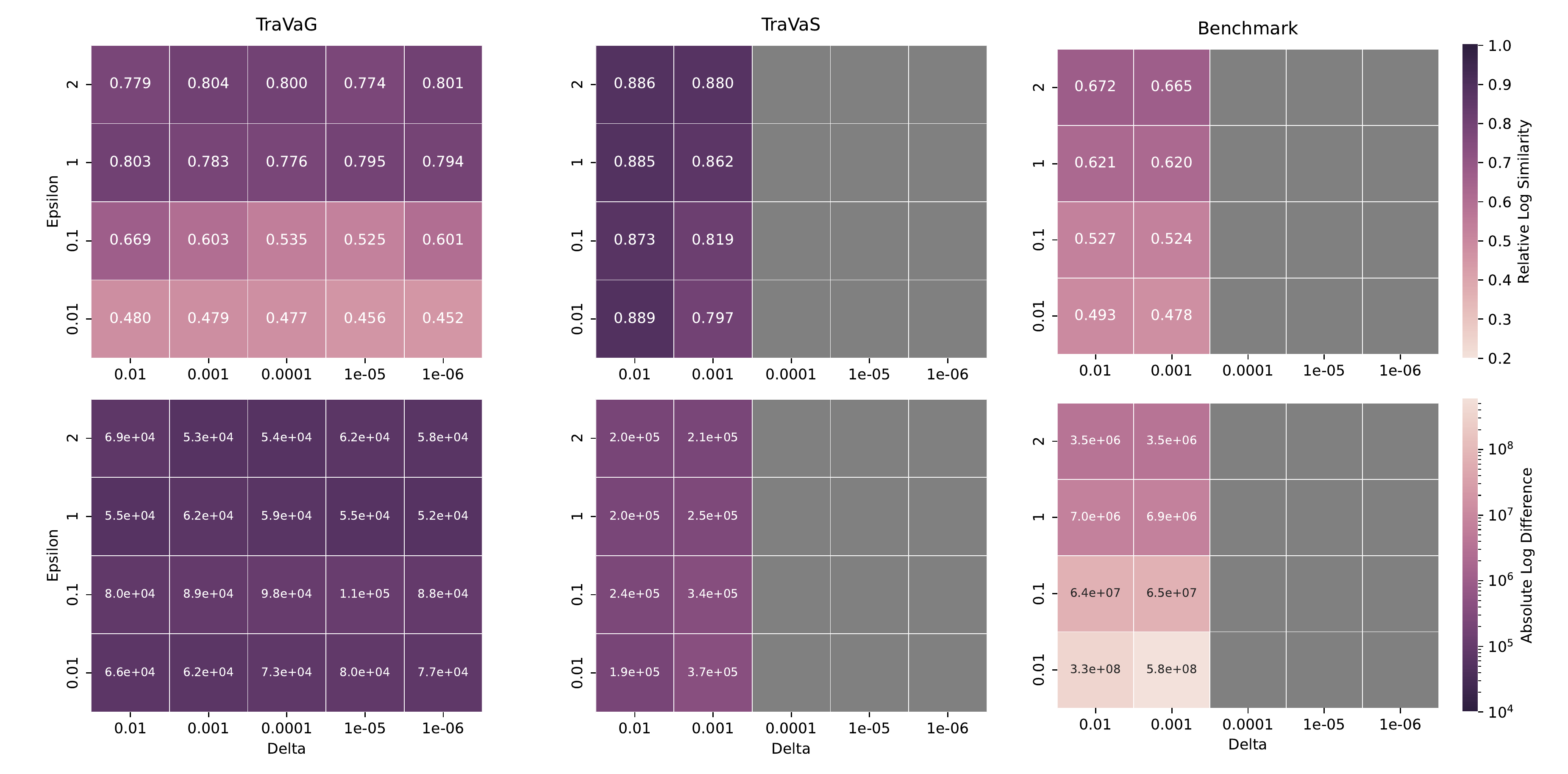}
\caption{The \emph{relative log similarity} and \emph{absolute log difference} results of anonymized BPIC13 logs generated by TraVaG, TraVaS, and the benchmark method. Each value represents the mean of 100 generations for TraVaG and 10 algorithm runs for TraVaS and the benchmark method.} \label{fig_exp1}
\vspace{-0.5cm}
\end{figure}
%\vspace{-3 mm}

%\vspace{-0.42cm}
\subsection{Data Utility Analysis}\label{subsec:exp_datautil}
%\vspace{-0.4cm}
In this subsection, the results of the two aforementioned data utility metrics are presented for both real-life event logs.
Figure~\ref{fig_exp1} shows the average results on BPIC13 in a six-fold heatmap. The gray fields at the TraVaS and benchmark methods denote an unsuccessful algorithm execution.
For $\delta<10^{-3}$, the thresholding of TraVaS becomes too strict and removes many variants in the anonymized outputs. On the contrary, the benchmark method introduces artificial variants and noise to an extent that is unfeasible to average within reasonable time and accuracy.
In opposition, TraVaG successfully manages to generate anonymized outputs for $\delta<10^{-3}$. More importantly, both results of \emph{relative log similarity} and \emph{absolute log difference} do not illustrate clear decreasing trends on lower $\delta$ within the investigated parameter range. We explain this expected observation by the fact that TraVaG avoids any pruning mechanism on its output and implements less $\delta$-dependent Gaussian noise via RDP into the gradients (see Section~\ref{subsec:privacy_acc} and \cite{rdp}).
% In direct competition to TraVaS and the benchmark, our generative model moreover outperforms in terms of absolute log difference and the benchmark with respect to relative log similarity (distribution overlap).

\begin{figure}[bt]
\centering
\includegraphics[width=0.92\textwidth,keepaspectratio]{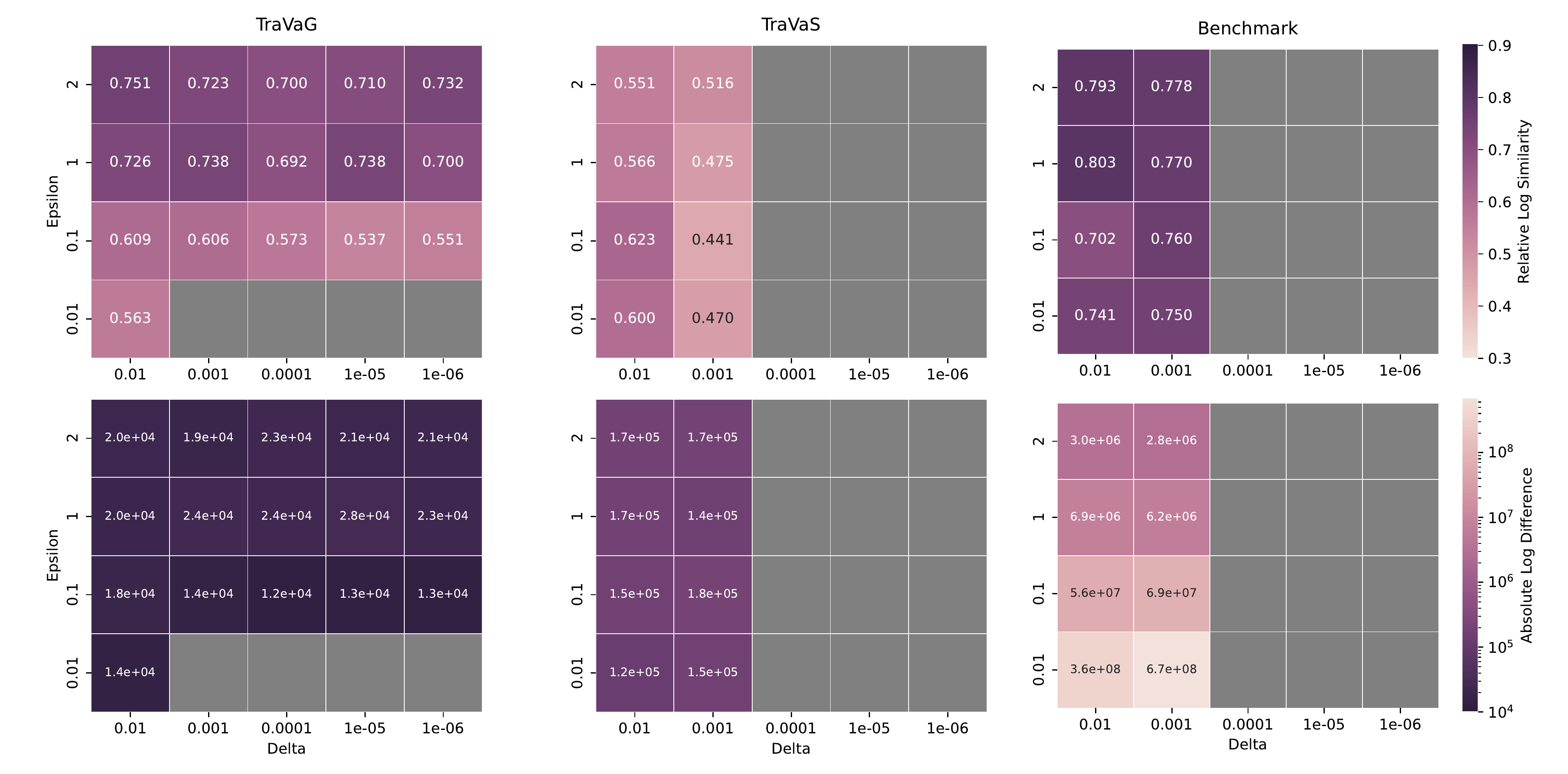}
\caption{The \emph{relative log similarity} and \emph{absolute log difference} results of anonymized Sepsis logs generated by TraVaG, TraVaS, and the benchmark method. Each value represents the mean of 100 generations for TraVaG and 10 algorithm runs for TraVaS and the benchmark method.} \label{fig_exp2}
\vspace{-0.5cm}
\end{figure}

Whereas the absolute log difference results maintain a rather stable output for the different ($\epsilon,\delta$) values, the TraVaG relative log similarity presents a strong positive $\epsilon$-dependency. As a result, the absolute statistics (absolute Levenshtein distances and absolute frequencies) of the anonymized event data seem to be more similar to the original logs as the variant distributions. A rationale for this discrepancy lies in the still comparably small dataset with 7554 instances over 1511 variants (features). By construction, TraVaG accomplishes reproducing equally sized event logs containing many original variants but fails to pick up some characteristics of the underlying distribution once the input data or the training iterations are limited. Hence, we expect this diverging trend to diminish with increasing training data.
%\vspace{-6 mm}

The data utility results for the Sepsis log are presented in Figure~\ref{fig_exp2}.
With only 1050 instances at 846 variants (features), this dataset is even smaller and thus more difficult to train for TraVaG than BPIC13.
As a result, we observe similar, but more pronounced behavior of relative log similarity and absolute log difference metrics compared to Figure~\ref{fig_exp1}. An extreme example are the results at $\epsilon=0.01, \delta<10^{-2}$, where the introduced gradient noise turned out as too intense for the generative model to converge under the given training data size.
For the remaining privacy settings, TraVaG again outperforms its competitors at the absolute log statistics while the relative log similarity performs slightly better than TraVaS and at the same order as the benchmark results for $\epsilon>0.1$. 
% In this context, we note that due to the high trace uniqueness on the Sepsis log, distribution-based utility metrics applied to TraVaS or the benchmark have to be treated with care (see experiments in \cite{rafiei_travas_short}).

\subsection{Process Discovery Analysis}\label{subsec:exp_procdisc}

\begin{figure}[bt]
\centering
\includegraphics[width=0.92\textwidth,keepaspectratio]{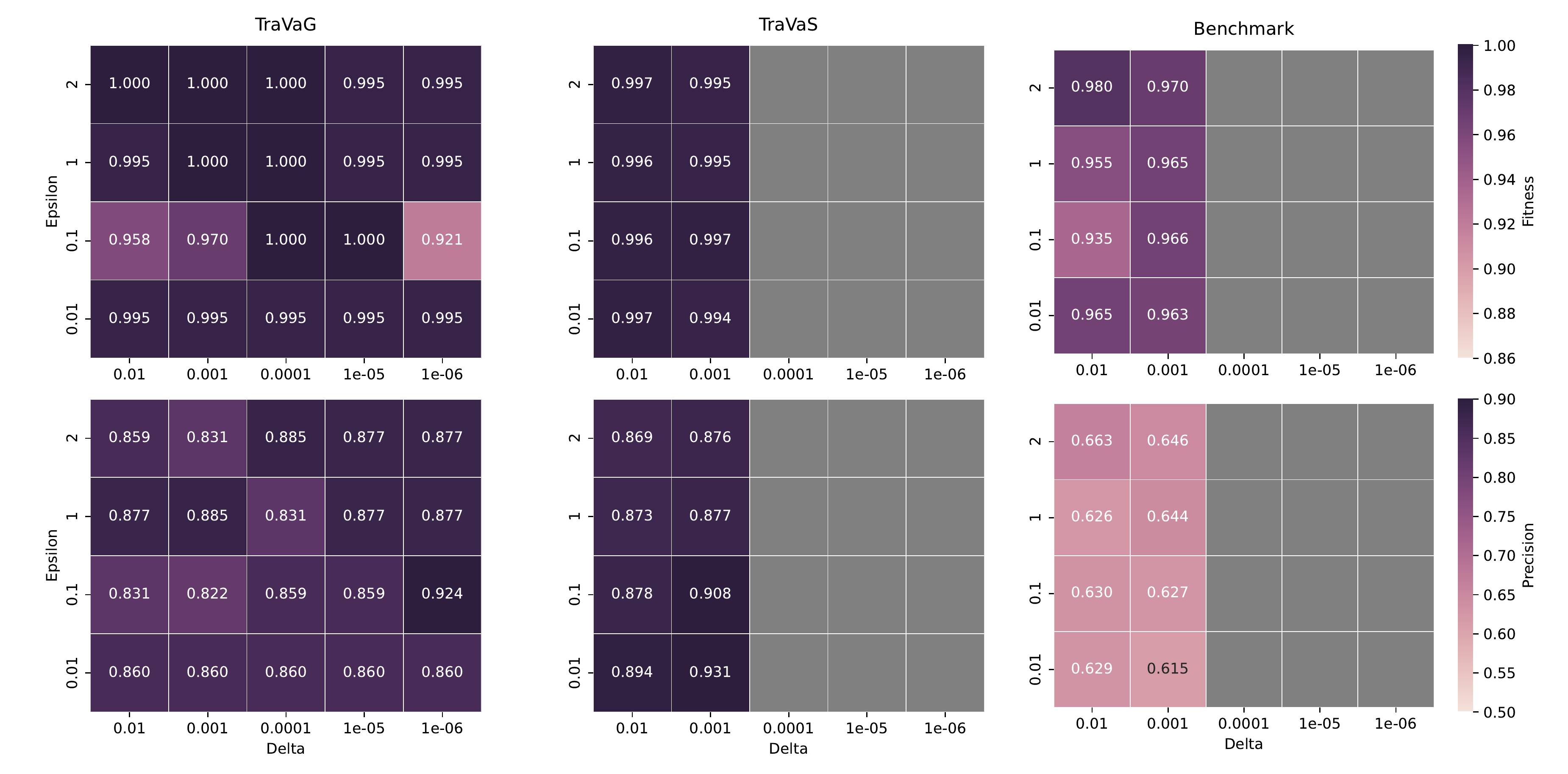}
\caption{The \textit{fitness} and \textit{precision} results of anonymized BPIC13 event logs generated by TraVaG, TraVaS, and the benchmark method. Each value represents the mean of 100 generations for TraVaG and 10 algorithm runs for TraVaS and the benchmark method.} \label{fig_exp3}
\vspace{-0.5cm}
\end{figure}

% Our data utility analysis is complemented with a process discovery evaluation based on \emph{fitness} and \emph{precision} scores.
Figure~\ref{fig_exp3} illustrates the result utility analysis of TraVaG, TraVaS, and the benchmark on BPIC13.
As discussed in Subsection~\ref{subsec:exp_datautil}, TraVaG successfully manages to produce results for $\delta<10^{-3}$ where the other methods are not applicable.
Except for the three outliers at $\epsilon = 0.1$, both fitness and precision show a stable distribution without considerable dependence on the different privacy parameters. In accordance with Figure~\ref{fig_exp1}, we thus conclude that the absolute log difference provides a better proxy for process-discovery-based performance of TraVaG than relative log similarity.
Similarly, the strong scores on both metrics demonstrate a sufficient replay behavior between the model obtained from an anonymized log and the original log. Whereas fitness denotes that the process model still captures most of the real underlying event data, precision depicts only a small fraction of model decisions, not being included in the anonymized event log.
Consequently, TraVaG accomplishes learning the most important facets of the BPIC13 variant distribution for the discovery algorithm to produce a fitted model.
When compared to the alternative methods, TraVaG achieves comparable scores as TraVaS and again outperforms the benchmark.

\begin{figure}[bt]
\centering
\includegraphics[width=0.92\textwidth]{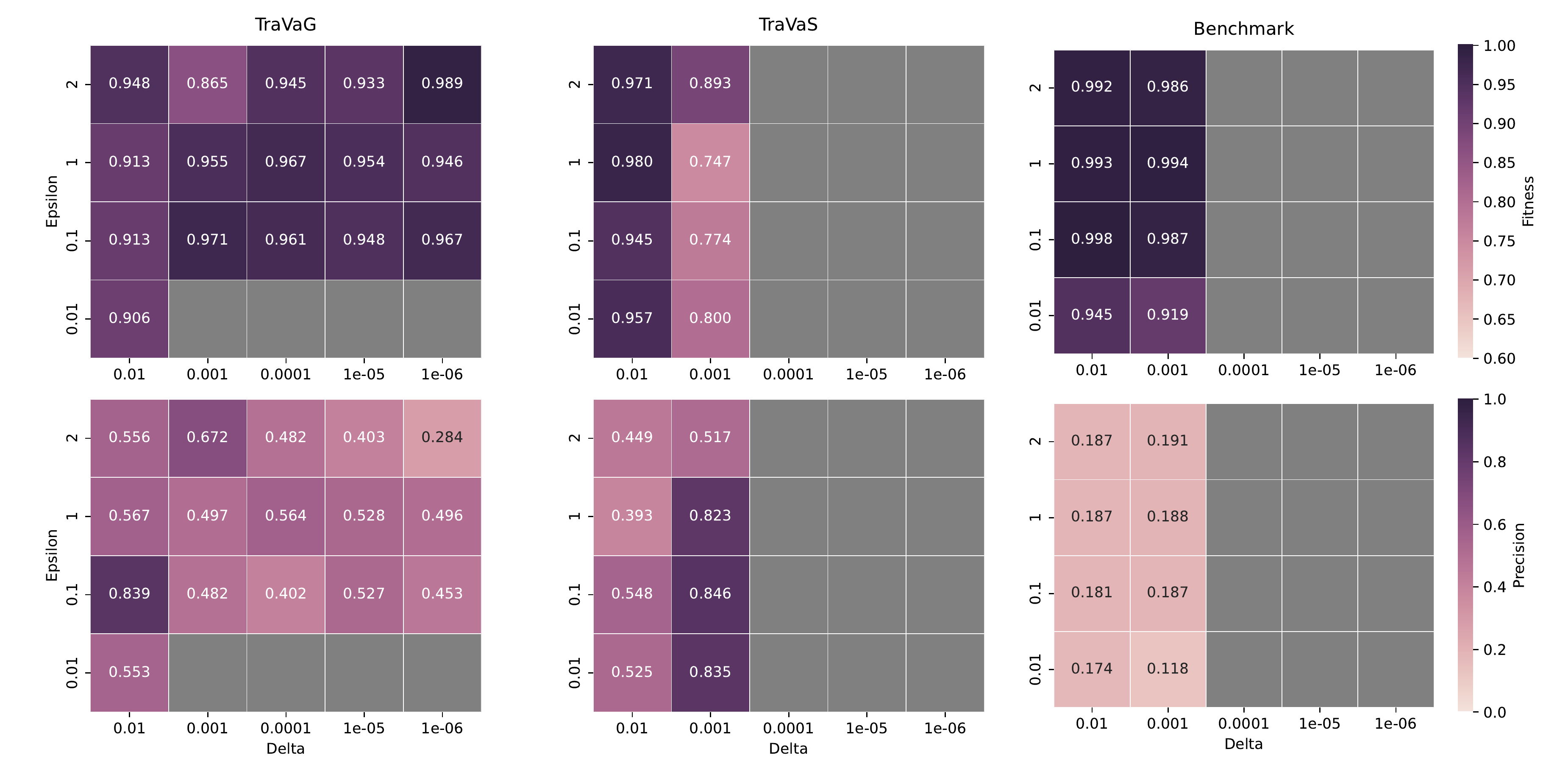}
\caption{The \textit{fitness} and \textit{precision} results of anonymized Sepsis event logs generated using TraVaG, TraVaS, and the benchmark method. Each value represents the mean of 100 generations for TraVaG and 10 algorithm runs for TraVaS and the benchmark method.} \label{fig_exp4}
\vspace{-0.5cm}
\end{figure}

The result utility evaluation of the high trace-unique Sepsis log is presented in Figure~\ref{fig_exp4}.
With respect to fitness, TraVaG shows similar values as TraVaS but a slight under-performance compared to the benchmark method. The main cause for this observation again refers to the infrequent variants and the small log size. While TraVaS maintains a strong $\delta$-related threshold and TraVaG copes with the limited training data, the benchmark method introduces many artificial variants but tends to match the frequent traces. As a result, the discovered process models are able to replay most of the original behavior in contrast to TraVaG and TraVaS results.
According to the aforementioned explanation, precision reflects an inverted trend.
Here, the larger models of the benchmark method contain many possible decision paths that are nonexistent in the underlying event log. For TraVaS and TraVaG, we thus achieve more precise anonymized process models.

% \begin{table}[h]
% \scriptsize
% \centering
% \caption{General statistics of the event logs used in our experiments.}
% \label{tab:exp_data}
% \begin{tabular}{c|c|c|c|c|c}
% \hline
% Event Log & \#Events & \#Cases & \#Activities & \#Variants & Trace Uniqueness\\
% \hline
% Sepsis & 15214 & 1050 & 16 & 846 & 80\%\\
% BPIC13 & 65533 & 7554 & 4 & 1511 & 20\%\\
% BPIC-2012-App & 60849 & 13087 & 10 & 17 & 0.12\%\\
% \hline
% \end{tabular}
% \label{tab:logs}
% \end{table}

\vspace{-0.15cm}
\section{Conclusion}\label{sec:conc}

TraVaG has shown that training a differentially private combination of autoencoders and GANs to synthesize anonymized event data from an underlying original variant distribution outperforms current state-of-the-art selection-based variant anonymization techniques and prefix-based approaches. Particularly, for strong privacy at the low $\delta$ range. 
Moreover, TraVaG has the unique advantages of outstanding resource-efficient execution, the absence of distorting noise thresholds, a general acceptance of continuous data streams, and no fake variant generation.
In combination, these characteristics allow TraVaG to efficiently operate with infrequent variant data in the low $\delta$ regime without real competitors. Nevertheless, we note that the framework comprises a more complex training procedure and privacy budget accounting than approaches that directly digest DP parameters, such as TraVaS \cite{rafiei_travas_short}.
% Due to the DP-SGD mechanisms that are based on RDP, the conventional DP parameters $(\epsilon, \delta)$ cannot be directly extracted from or inserted into the TraVaG model \cite{rdp}.
We have to follow the one-way procedure to first obtain RDP parameters $(\epsilon, \alpha)$ from noise multiplier $\Phi$, sampling rate $q$, iterations $T$ and then convert $(\epsilon, \alpha)$ to $(\epsilon, \delta)$. Note that a similar procedure is followed by other techniques that are based on RDP, such as Libra \cite{libra}.
Consequently, specific privacy levels can only be ensured by repeatedly analyzing different TraVaG network settings until a successful match is found. This hyperparameter dependence could be studied in more detail and even coupled with a fully automated tuning strategy in future work. 

\bibliographystyle{splncs04}
\bibliography{references}

\end{document}